# Linguistic Taboos and Euphemisms in Nepali

Nobal B. Niraula
Nowa Lab
Madison, Alabama, USA
nobal@nowalab.com

Saurab Dulal
The University of Memphis
Memphis, Tennessee, USA
sdulal@memphis.edu

Diwa Koirala
Nowa Lab
Madison, Alabama, USA
diwa@nowalab.com

## ABSTRACT

Languages across the world have words, phrases, and behaviors - the taboos - that are avoided in public communication considering them as obscene or disturbing to the social, religious, and ethical values of society. However, people deliberately use these linguistic taboos and other language constructs to make hurtful, derogatory, and obscene comments. It is nearly impossible to construct a universal set of offensive or taboo terms because offensiveness is determined entirely by different factors such as socio-physical setting, speaker-listener relationship, and word choices. In this paper, we present a detailed corpus-based study of offensive language in Nepali. We identify and describe more than 18 different categories of linguistic offenses including politics, religion, race, and sex. We discuss 12 common euphemisms such as synonym, metaphor and circumlocution. In addition, we introduce a manually constructed data set of over 1000 offensive and taboo terms popular among contemporary speakers. This in-depth study of offensive language and resource will provide a foundation for several downstream tasks such as offensive language detection and language learning.

## 1 INTRODUCTION

**Disclaimer**: Due to the nature of this topic, this paper contains highly offensive language, hate speech and adult content. They don't reflect the views of the authors in any way. The goal of this paper is to build a foundation to stop hate speech and discrimination in the use of language.

Offensiveness is a socio-cultural and situational phenomenon. Constructing a universal set of offensive terms for all the languages is near impossible because a culture, language, or situation defines what terms, phrases, or actions are offensive. Furthermore, the semantics of terms keeps changing with time. These variations warrant the language and culture-specific study of taboos and offenses[1]. This paper presents the study of linguistic taboos and euphemisms in Nepali, a language belonging to the Indo-Aryan family. Nepali is the official language of Nepal and several eastern Indian regions including Sikkim, Darjeeling, and Kalimpong. It is spoken by more than 20 million people, mainly in Nepal, and other places in the world including Bhutan, India and Myanmar[2]. To the best of our knowledge, this is the first study of linguistic taboos and euphemisms in Nepali. Thus, it provides a foundation for many Natural Language Processing (NLP) tasks such as offensive language detection, comment moderation and natural language generation.

Today social media, online forums and news portals are increasingly under pressure to stop offensive content. Manually reviewing streams of content is labor-intensive, costly and slow. NLP can greatly assist the moderation task. However, NLP for this task requires a language-specific understanding of offensive constituents. To this end, this study provides great insights into understanding how offensive languages are formed, what varieties of offensive content exist and how to generate non-offensive content in Nepali. Furthermore, it also provides benefits to Nepali language learners e.g. to avoid making offensive content and apply euphemisms to circumvent embarrassing situations.

Language is a medium through which one expresses thoughts and feelings to others. It is used in every aspect of human communication, expressing emotions such as love, hate and anger, revealing identity, disseminating information, and so on. Although language is used freely, society sometimes imposes restrictions and constraints on the use of certain terms, expressions, or behaviors, called "taboos". Breaking a taboo is usually considered objectionable because it is harmful to the social, religious, or ethical values of the society. Taboo can be verbal or non-verbal. The non-verbal taboo (aka. behavioral taboo) restricts the use of behavioral patterns that are offensive to the social values or customs. For instance, *entering a house with shoe on*, and *eating food during a classroom session* are non-verbal taboos in Nepali culture. The verbal taboo (aka. linguistic taboo) restricts the use of certain words or parts of a language in public discourse due to social constraints e.g. *kando* (*buttock*). Therefore, instead of using linguistic taboos, euphemisms are used to alleviate or erase the harshness. They are roundabout ways of expressing unpleasant terms to neutralize or lessen the unpleasantness.

Nonetheless, people do deliberately use linguistic taboos to make hurtful, derogatory, and obscene comments. However, a language does not necessarily need linguistic taboos to be offensive. Some words or phrases that do not have taboo meaning lexically can also be offensive if used in certain contexts. For example, the sentence *tan kukur hos* (you are a dog) does not have any unmentionable linguistic taboo, yet it is an offensive sentence. In this study, we refer to both linguistic and contextual taboo terms as offensive terms.

---

[1] Throughout this paper, the word "offense" is explicitly used to refer linguistic offense only

Authors' addresses: Nobal B. Niraula, Nowa Lab, Madison, Alabama, USA, nobal@nowalab.com; Saurab Dulal, The University of Memphis, Memphis, Tennessee, USA, sdulal@memphis.edu; Diwa Koirala, Nowa Lab, Madison, Alabama, USA, diwa@nowalab.com.

[2] https://en.wikipedia.org/wiki/Nepali_language

The offensiveness is determined entirely by pragmatics[3] variables such as socio-physical setting, speaker-listener relationship along with the words used and tone of voice [14]. Many actions or expressions that are considered normal in one society may not be considered normal in other societies [15]. For example, asking person's age can be considered normal in Nepali culture due to its honorific system. People use appropriate pronouns and verbs in conversations based on the relative age. A younger person uses pronouns like *tapai, hajur* when talking to an elder, the latter uses pronouns like *timi, tan* to the younger. In contrast, asking person's age isn't considered normal in Western culture [10]. Similarly, touching a book with feet is fine in Western society while it is examined as sinful act in Nepali culture. Likewise, diverse animal names are used for personal insults in Nepali (more on Sections 3 & 4) compared to English which has fewer cases like *bitch, pig, ass, snake, and monkey*.

Different dialects of a language can have extremely different meanings of a term. The term *coger*, for example, is a normal word in Colombian Spanish which means *to take* but it is a taboo word in Argentine Spanish which refers *to have sex*. Similarly, *Chholnu* is a normal word in Far-Western Nepal and means *to scrub*. This is a taboo term in Eastern Nepal and means *to masturbate*. Offensiveness level can also be different across the dialects e.g. *bloody* in Australian and American English is not as offensive as it is in British English [22, 27].

The offensiveness of a term evolves with society. The term *breast meat* was considered offensive in the $20^{th}$ century and *white meat* was its euphemism [10]. It has lost the offensiveness in modern-day English. Similarly, in Nepali, the term *bhalu* had just a single meaning in the past referring to the animal *bear*. Nowadays it can also mean a *prostitute* in an offensive context.

Ambiguity is an inherent property of a language. Pragmatics and contextual evaluation are required to measure the offensiveness of a term. The word *gadha* (donkey) is offensive in the sentence *ta gadha hos* (you are a donkey) but is not offensive in *gadhale bhari bokchha* (donkey carries load). Likewise, the word *baby* is more offensive to young boys than to their parents [15].

In this paper, we present a detailed corpus-based study of offensive language in contemporary Nepali. We collected and analyzed thousands of offensive social media posts and comments from diverse sources. Specifically, we studied over 7000 social media posts and manually collected over 1000 tabooed and offensive terms. Furthermore, we identified 18 different varieties of linguistic offenses such as Bodily Excretion, Death, Disease, Relation, Sex, Gender, Plant, and Animal. Our major research contributions are as follows:

(1) The detailed study of linguistic offenses in 18 different categories
(2) A reference data set of over 1000 linguistic taboos and their categories
(3) Common euphemism techniques

## 2 BACKGROUND

In this section, we provide a general overview of the taboo and offensive language and related work.

### 2.1 Taboo Language

Wardhaugh defines taboo as the things that are prohibited or avoided in any society of behavior because they are believed to be harmful to its members and would cause them anxiety, embarrassment, or shame [29]. If necessary, they are expressed in very roundabout ways, i.e., using euphemisms. Popular English dictionaries also define taboo similarly: Cambridge Dictionary of English[4] defines *taboo* as a subject, word, or action that is avoided for religious or social reasons. According to Longman Dictionary[5], a taboo subject, word, activity, etc. is one that people avoid because it is extremely offensive or embarrassing. It is not accepted as socially correct or it is too holy or evil to be touched or used.

The word *taboo* is derived from the Tongan word *tabu* towards the end of the eighteenth century [1]. Its first use was found in the log entry for 15 June 1777 by Captain James Cook who was sent to Tahiti. In Togan, the languages of Polynesia, the word *tabu* was used to reference things that are forbidden, holy, or untouchable. At present, there are wide varieties of tabooed subjects including sex, death, illness, excretion, bodily functions, and religious matters [14, 29].

There exists numerous sociolinguistic studies of taboos for several languages but Nepali. Gao [10] investigated English taboos and euphemisms. She described seven different varieties of taboos in English: bodily excretions, death and disease, sex, four-letter words, swear words, privacy, and discriminatory language. Similar studies also exist in Igbo society in Nigeria [9], Algerian Society [11], Yemani Culture [21], Karonese Culture in Indonesia [3], Russian [12] and Finnish [13].

Closest to our work is by Behzad et al. who studied linguistic taboos in Pahari culture, the culture of the hilly northern part of India, Pakistan, and Nepal [4]. Pahari language is the mother tongue of people of Azad Jammu and Kashmir where the majority of people are Muslims. Since Hindu is the majority in Nepal and Nepali is the most spoken language, the study of linguistic taboos in Pahari does not apply to Nepali language and culture.

### 2.2 Offensive Language

Offensive language is used to make hurtful, derogatory, and obscene comments to a target – usually an individual, organization, or a social group. As mentioned previously, offensive language can contain unmentionable as well as contextual linguistic taboos.

Nowadays people heavily use computer-mediated communication (CMC) such as social media, emails, and blogs to express opinions and exchange messages. CMC differs in many ways with traditional communication. Users can partially or fully anonymize their identities behind the Internet. The anonymity provides strength to express things that one cannot express in someone's face [24]. With anonymity and physically distant from their audiences, offensive language is more likely to occur on the Internet and has a far-reaching and determinative impacts [24]. The proliferation of hate speech in CMC has created a huge research interest and momentum towards automated offense detection in social media such as identifying offensive posts, comments, and memes [23].

---

[3]Branch of linguistics that deals with context of utterance [16]

[4]https://dictionary.cambridge.org/us/dictionary/english/taboo
[5]https://www.ldoceonline.com/dictionary/taboo

| |
|---|
| Nepali (Devanagari): मासाला पागल भए जस्तो छ! <br> English: it seems he got *pagal* (mad) |
| Nepali (Transliterated): *muji* talai ta ma *thookxuu gooli saala des drohi* <br> English: *muji* (pubic hair) I will shoot you, traitor |
| Nepali (Transliterated): *sale khate* aphu matra educated thhanndo rahexa <br> English: *sale* (brother-in-law) *khate* (pejorative term for people living in urban slum dwellers) thinks he is the only educated |
| Nepali (Transliterated): *bahun* le desh bigare <br> English: *bahun* (an upper cast) spoiled the country |

Table 1: Sample Nepali social media posts and their English translations with offensive terms in Italics. Posts are written in Devanagari as well as in Transliterated Roman Script

To the extent of our knowledge, there is not a single study of identifying offensive content in the Nepali language. In other languages, this has been studied under different topics including detection of hate speech [7, 8], abusive language [19, 20], cyberbullying [5, 25, 30], aggression [2, 6], and insults [26]. Recently, shared task initiatives such as SemEval Task 6 [32] for English and GermEval [31] for German have united researchers to develop and evaluate their approaches in consistent ways. These efforts have produced resources such as training and testing data sets that are critical for identifying offensive content and their types and targets.

Although the goal of this paper is to study and characterize offensive language, taboos and euphemisms in Nepali, the data set of tabooed and contextually offensive terms that we have produced can serve as a baseline system for offense detection. Such system can flag a content in CMC (e.g. a social media post) as offensive if any of the terms in the data set is present in the content.

We follow a corpus-based study in which we investigate a data set of human conversation to identify tabooed and offensive terms. Social media such as Twitter[6], YouTube[7], Facebook[8] and Blogs are excellent sources of human conversation where users discuss a wide range of topics including politics, gender, religion, movies and television shows, and so on. They participate in the discussions by writing, sharing, reacting, and commenting. Some users in these conversations express their opinions which are insulting or offensive to others. These offensive content are of our interest as they contain tabooed and offensive terms or phrases.

## 3 METHODOLOGY

Over 90% of current Nepali social media users are young (age < 54 years) [9]. With the older generation being more mature, we assume they generate very few offensive comments. The offensive terms that we identify from the current social media posts are the contemporary offensive terms mostly popular among the younger generation.

### 3.1 Corpus to Study Offensive Terms

Social media provides a platform for people to discuss in public, private, or secret forums. We are particularly interested in the conversations at public forums because any offensive language used in such places is not acceptable to the general audience. Since most of the posts in social media are non-offensive (e.g. around 0.2% swear word rate for Myspace [14]), it is highly likely that we obtain just a few offensive posts even while annotating a very large number of posts. So, we narrowed down the search by manually listing posts presumably rich in offensive content. The identified posts were either outright offensive (e.g. vulgar YouTube videos, sexist and racist Tweets), or had potentially offensive content (e.g. racial hatred posts), or had generated debate among participants (e.g. political issues). In addition to the social media posts, we also extracted the associated comments. In total, we obtained over 7000 records including both the posts and the comments.

Sample social media posts that contained offensive language are shown in Table 1 along with the corresponding English translations. The offensive terms in the original Nepali are highlighted in italics. If the translation seems to undermine the offensiveness level, we kept the offensive term as is in the English translation. Note that the last example in the table *"bahun le desh bigare"* does not contain a single offensive word but the overall meaning of this post is offensive i.e. racist comment targeting a cast, *bahun*.

### 3.2 Data Set Creation

Three human annotators, all native Nepali speakers, studied the social media posts and identified the offensive terms. Confusing and borderline terms were discussed together and the final decision was taken in consensus. As expected, the social media posts were very noisy and contained many variants of the same term. For example, the offensive term *khate* was written as *khatee*, *khaate*, *khte*, *khatey* etc. Besides, the social media posts were written in Devanagari script and Romanized text (using phonetics) for the ease of typing. We extracted all of these raw variants, normalized them, and included in the data set.

We normalized the raw offensive terms and obtained their synonyms using Nepali Shabdakosh[10] dictionary service. However, since the offensiveness of synonyms can be different from its term's offensiveness, manual revision was required. For instance, *pisap* is a non-offensive synonym of the offensive term *mut* (urine). In total, we discovered 1077 offensive terms and made them publicly available[11]. We believe this first public data set of linguistic taboos for a low-resource Nepali language can facilitate the research towards this important topic.

---

[6] http://twitter.com  
[7] http://youtube.com  
[8] http://facebook.com  
[9] https://www.slideshare.net/DataReportal/digital-2019-nepal-january-2019-v01  
[10] http://nepalishabdakosh.com  
[11] https://github.com/nowalab/offensive-nepali

| RawRom | RawNep | NormNep | NormRom | IsTaboo | Class |
|---|---|---|---|---|---|
| kandoo | कन्ढो | कन्डो | kando | 1 | Body Part |
| boksi | वोक्सी | बोक्सी | boksi | 1 | Others |
| goru | गोरू | गोरु | goru | 0 | Animal |
| damini | दमीनी | दमिनी | damini | 1 | Race |
| pani pade | पानी पादे | पानी पादे | pani pade | 0 | Excretion |
| aatankabadhi | आतङ्कवादी | आतङ्कवादी | aatankabadi | 0 | Politics |
| goli hanna | गोली हान्न | गोली हान्नु | goli hannu | 0 | Action |

Table 2: A snippet of the manually constructed data set. Notations: Rom=Romanized, Nep=Nepali and Norm=Normalized. IsTaboo column value 1 indicates if the term is a taboo term.

The snippet of the data set is shown in Table 2. It consists of *RawRom* and *RawNep* designating raw terms in Romanized form and in Devanagari script respectively. Their normalized forms are stored at *NormNep* and *NormRom* columns. In addition, we labeled "1" for taboo terms and "0" for contextually taboo terms in *IsTaboo* column. We also provided a primary class type (more on Section 4) for each entry in *Class* column.

The statistics of the manually derived data set are presented in Table 3. Out of 1077 identified offensive terms, 158 (about 15%) were absolute taboo, offensive outright. Body Part, Race, Relation, Sex, and Vulgar are the top categories that contained a higher proportion of absolute taboos. Categories such as Action, Cloth, Animal, Death, Geographic Location, Plant, Politics, Religion, and Weapon have very few or no taboo terms at all. Thus, the terms in these categories require contexts to be offensive. Interestingly, Politics, with 115 offensive terms, is the largest category. Race with 77 offensive terms is the second-largest category. Unlike in English, use of diverse animal names are observed in Nepali offensive language. With 72 terms, Animal category is the third-largest category in our data set. As expected, universal taboo topics such as Sex, Vulgur, and Body Parts are also in the top most popular categories. Lastly, relations being very important in Nepali culture, the offensive terms related to kinship are also very common – a unique characteristic of Nepali culture compared to the Western cultures.

## 4 VARIETIES OF LINGUISTIC OFFENSES IN NEPALI

We carefully studied the data set of offensive terms that we constructed in Section 3 and identified over eighteen interesting varieties (categories) such as Animal and Plant names, Race, and Sex. While some of these categories exist in other cultures and languages, others are unique to Nepali culture. Besides, the identified varieties are not mutually exclusive, i.e, a term can belong to multiple categories. Importantly, the main rationale for the selection of these categories is to explore several dimensions of linguistic offenses existing in the language, keeping in mind that these categories will largely assist to do a fine-grained classification in offense delectation task. In this section, we describe these categories in detail.

**1. Action**: We identified many offensive comments that contained action verbs or expressions that were threatening, defaming, or described or called for harmful action to the target (e.g. a person or an organization). Examples:

- Yeslai chhitai goli **thokna** parcha (Need to shoot him soon)
- Party karyaalaya **aaago lagauna** jaau (Let's go to burn down the party office)

Other actions include *katnu, pitnu, jalaunu, pata kasnu, lakhetnu, jel halnu, sutnu, chikhnu, mutnu, chusnu, polnu, chhiraunu, latta hannu* and so on. Most of the actions are context-sensitive i.e. offensiveness depends on their use in the language.

**2. Body Part**: Body parts are commonly tabooed across different cultures and languages. Female and male genital parts and pubic hair are tabooed outright while others need contexts to be tabooed. Our observation shows that body parts mostly appear in comments that are vulgar, racist, sexist, and personal insults. Commonly mentioned body parts include *puti, gula, chak, laando, geda, laantho,*

| Category | Taboo | Contextual | Total |
|---|---|---|---|
| Action | 0 | 45 | 45 |
| Animal | 0 | 72 | 72 |
| Body Part | 21 | 23 | 44 |
| Cloth | 0 | 12 | 12 |
| Death | 1 | 13 | 14 |
| Disability | 6 | 31 | 37 |
| Disease | 3 | 25 | 28 |
| Excretion | 4 | 17 | 21 |
| Gender | 7 | 8 | 15 |
| Geographic Location | 0 | 10 | 10 |
| Idiom | 3 | 26 | 29 |
| Marriage | 15 | 15 | 30 |
| Plant | 0 | 12 | 12 |
| Politics | 0 | 115 | 115 |
| Race | 12 | 65 | 77 |
| Relation | 11 | 32 | 43 |
| Religion | 0 | 12 | 12 |
| Vulgar | 31 | 19 | 50 |
| Weapon | 0 | 17 | 17 |
| Other | 44 | 350 | 394 |
| **Grand Total** | **158** | **919** | **1077** |

Table 3: Summary of offensive data set manually mined from thousands of social media posts

*poka, thun, thun ko tuppo, junga, jhus, naak, kapal, taauko, muji*. Examples:

- Sabaiko **taauko** katera gift pathaidim (Let's behead everyone and send the heads as a gift)
- **Muji** tero kaam chaina, turuntai rajinama de (*Muji* you are useless, resign immediately)

Interestingly, the expressions following the pattern: (tera | tero | teri | mero | meri) + (baau | ama | hajurba | budi) + ko + body part (especially head–*taauko* or genital parts) are offensive, e.g., *tero baau ko taauko* (your father's head), *teri amako puti* (your mother's vagina), *tero hajurba ko taauko* (your grandfather's head).

Because body parts are often required to be mentioned in normal conversations, euphemisms related to body parts are widely used in the Nepali. *Wakṣhyastal and chhati* are used for breast, *guptaanga* is used for male or female genitals, *yoni* is used for vagina, and so on. For instance, *mahinawari bhako bela "yoni" safa rakhnu jaruri cha* (it is important to sanitize your vagina during menstruation). Interestingly, the term *linga* has multiple senses which can mean *penis* (a taboo term) or the word gender (non-taboo term). *Shiva Linga* (Lord Shiva's penis), however, is a non-taboo term which is a revered icon in Shiva temple.

**3. Cloth:** Clothes that have direct connections with the internal body parts or sexual organs are contextually tabooed in Nepali language. Words such as *thun choli, bra, petikot, taalo – sanitary pads, kanchuki, kachha, kattu* etc. are commonly avoided in public and euphemisms are used in substitution. *Aaimai ko bhitri bastra* or simply *bhitri cholo* is used instead of *bra* and *thun choli*. Similarly, *bhitra lagaune* is used for *kachha, kattu, kanchuki* etc. The use of alternatives are also contextual, for example *bhitra lagaune* can refer to underwear, but it can also refer to *bra* or *petikot* in a separate context. Example: *mero bhitra lagaune khoi* (where is my underwear).

Additionally, we identified clothes that are unrelated to internal body parts or sexual organs are contextually used in offensive languages. Words such as *seto kapada or katro* (shroud) are used to express personal attacks: *talai katro odayera ghat ma purauchu*. *Dhoti, topi, kandani, janai*, etc are used to express racial offenses. Examples:

- **Dhoti** lai india tira lakhetna parcha (Should chase this *dhoti* out towards india)
- **Kandane (or toppe)** [12] bahun

**4. Animal:** Animals are the commonly used metaphors in Nepali offense. Frequently used creatures in offensive dialect include *badar* (monkey), *gadha* (donkey), *kukur* (dog), *bheda* (sheep), *hanuman* – a divine monkey in Hinduism – is used here to refer a fanatic follower, *goru* (ox), *bwaso* (hyena), *bwasha, bhaishi* (buffalo), *hatti* (elephant), *boka* (he-goat), *bakhri* (she-goat) and *pothi* (hen). Most of these names are used in general offence, however, we also found some of them being used in sexist and racial comments. Examples:

- **Boka** haru **bakhri** khojna hide (*Boka* – pervert boys are heading in search of *bakhri*)
- Ta pani tah tai party ko **hanuman** hos ni (You are also a *hanuman* of the same party)
- Anuhar nai **bhaisi** ko jasto cha (Face looks like that of *buffalo*) (can be racist, depends on context)
- **Pothi** baseko suhaudaina (It does not suit a *woman* to raise her voice) (sexist idiom)

Interestingly, based on our observation, we found that the terms referring to female are more offensive than their male counterparts. For instance, *kukur* (dog) and *bandar* (monkey) are not offensive but their female versions *Kukurni* (bitch) and *Badarni* are taboo terms. This phenomenon is also observed in other cases e.g. (*newar, newarni*), (*rai, raini*) and (*chor, chorni*) – here, the masculine terms are generally non-offensive while the respective feminine terms are highly offensive.

**5. Plant:** Like animal names, plant names are also commonly used in offensive languages. Commonly used plant names include *kera* (banana), *aalu* (potato), *mula* (radish), *pharsi* (pumpkin), *kandamul* (wild plants), and *paat* (leaf). Examples:

- Mero **kera** khanchas? (Want to eat my penis?) (*kera* – banana is used here to indicated *penis*)
- Tyo **mula** ko kura nagar (Don't talk about that *mula* – radish)
- **Pharsi** ko *poi* (Husband of a fat women) (*pharsi* – pumpkin here means fat)

**6. Death and Disease:** Many words related to death and disease are contextually offensive and are frequently found in personal attacks. Examples include words such as *pagal, cancer, haija* related to diseases, and words such as *hariyo bash, katro, chihan, masan, maranchayshe, murda* related to death. Examples:

- Tero sarir le **hariyo baas** khojya jasto cha (Looks like your body is looking for a *coffin*) (context sensitive personal attack)
- Yo **pagal** ho (He is a *lunatic* – direct offense)

Euphemisms related to death include *bitnu, swargabash hunu, paralok hunu*, and *khasnu* for *marnu* (die) and *ghat* for *masan ghat* (cemetery). Disease related euphemisms include *manasik santulan thik nabhayeko* for *pagal* (mentally ill person) and a generic term *birami* (sick) for all kind of diseases.

**7. Disability:** Disability is another category that accumulates words often tabooed across different cultures and languages. Disability terms such as *lulo, langado, kano, bahiro, andho, psycho, kupro, lathebro, bhakbhake, maranchyashe* are tabooed when associated with human. They are not tabooed when associated with non-human entities e.g. bahiro sarkar (deaf government). Human rights and political movements have made many of these terms tabooed in recent years. Despite this, they are frequently used in offensive languages for personal insults and attacks. Examples:

- **Bahiro** chas ki kya ho? (Are you *deaf* or what?)
- Kulangar **saape** [13] (You *black-sheep* lame) – extremely abusive phrase referring to a physically disable person.
- Dekhdai khana na pako **maranchyashe** chas, k furti lagauchas. (You look *emaciated*, why are you showing off?)
- **lati** chhilli, *lato* jaat janai hali (*Dumb* pervert, showing off her dumb caste)

---

[12] *kandane* and *toppe* are terms especially used to abuse Bahun or bhramin cast. These terms come from the word *kandani* – long lace or belt to hold loin-cloth – and *topi* – hat respectively

[13] *Saape* comes from the word *saanp* which means snake. It is offensively used to indicate a handicap person abusing their inability to walk

Euphemisms are preferred when these terms are used for human. For example, *euta aakha na bhako, kaan nasunne, manasik santulan thik na bhako* are used instead of *andho, bahiro*, and *pagal* respectively.

**8. Excretions**: Most of bodily excretions such as *paad* (fart), *muut* (urine), *gu* (shit), *cher* (watery diarrhea), and *virya* (sperm) are tabooed. However, some others such as *ashu* (tears), *pasina* (sweat), *khakar* (cough), *dakar* (burp) and *ulti* (vomit) are not. Verbs associated with the production of the tabooed bodily excretions are also tabooed e.g. *padhnu* (to fart) for *paad* (fart) and *hagnu* (excrete) for *gu* (shit). Examples:

- Kasto **cher** jasto aachar (This pickle looks like a *shit*)
- Ta **gu** ko poka hos (You are a pack of *shit*)
- **Gobar** bhariyeko tero dimag (Your mind is full of *dung*)

Euphemisms for tabooed excretions are commonly used in public and hospitals e.g. *hawa* instead of *paad*, *aaka* instead of *gu / cher*, *susu* instead of *mut* and *seto pani* instead of *birya / shukrakit*.

**9. Marriage**: Marriage is one of the most prominent traditions in Nepalese culture. There are several rituals related to marriage. These rituals along with the associated human participants are identified using specific terms such as *dulaha* (bridegroom), *dulahi* (bride), *janti* (participants from bridegroom side), *daijo* (dowry), *kanya* (unmarried young girl or damsel). These terms can appear in the offensive comments related to the same. Traditionally, divorce was a rare phenomenon in Nepali culture, and women seldom remarry after the death of the husband. Living together before marriage is still a taboo. These activities resulted in many linguistic taboos that are unique to Nepali culture compared to western. Examples include: *bidhuwa* (widow), *bidhur* (widower), *batashe* (a child whose father is unknown), *rando* (). One interesting taboo expression is *baau ko bihe*, which means father's marriage. The expression is used to threaten the target with the harm that is so unprecedented that it will be equivalent to seeing his (target's) father's marriage, a very rare event to perceive in Nepali culture. Examples:

- Bhaisi lai daijo ko lagi **bihe** garya hola (Married a *bhaisi* for dowry) (*Bhaisi* i.e. buffalo is used here as a metaphor to indicate a fat/dark girl)
- Kunai **bidhawa** ko aakha nalagos (May a *widow's* curse not affect you)
- **Batashe** dherai nabol (Don't you talk more, *bastard*)

**10. Politics**: Politics has been a long standing subject around the world for centuries. The advent of social media and the connectivity among the people have fostered the expression of frustrations and disagreements in several topics regarding policies, political parties and their activities. In Nepali language, we found the following words contextually used offensively in political conversations: *desdrohi* (traitor), *bhrasta* (corrupt), *bhastrachar* (corruption), *mandale* (support of Nepali *panchayat*), *maobadi* (maoist), *rastraghati* (traitor), *bikhandankari* (separatist), *hatyara* (killer), *puchhar* (tail to refer a blind supporter), *fascist*, *bheda* (sheep – refer to fanatic follower), *bhate*, and *hanuman*.

The words in this category are temporal and can change with the dynamics of politics. A word that was neutral in the past may now be offensive and may become obsolete in the future. *Hanuman*, for example, had a single positive meaning in the past i.e. exemplifying the devotion towards his master, lord Ram. Currently, this word is also used to refer a fanatic follower of a party/ideology in an offensive context. Examples:

- **Deshdrohi** lai goli thok sala (Shoot the *traitor sala*)
- Ram ko bachan jharna pa chaina, **hanuman** haru le chati pitna thali sake (*Fanatic followers* started obeying their master in no time.)

**11. Vulgar**: Words that are explicitly and offensively used to refer sex, sexual acts, and human biological process are categorised as vulgar. People usually tends to avoid vulgar words, specifically sex talks and sexuality (menstruation, masturbation, and intercourse) in public as they are likely to cause anxiety. In Nepali too, almost all of the words that fall under this category are tabooed. Examples include: *chiknu* (have sex), *{gula, lado}* (penis), *puti* (vagina), *jhatha* (pubic hair), *phusi* (semen), *thankanu* (to have an erection). In addition, imported terms from English such *fuck*, *condom*, *cock*, *suck* and *blue film* are tabooed too. These terms are used very often in offensive remarks. Examples:

- Tero aama lai **chickne** mai hu bujhis? (I am the one who fucked your mother, do you understand?)
- **Lado** kha (Eat my *penis*)

As discussed above, the direct use of these words are eschewed and euphemisms are used instead. For example, *garba nirodhak sadhan* for *condom*, *swapnadosh* for *wet dream*, *purus or mahila ko guptaanga* for *lado* or *puti* respectively, *seto pani* for *briya* and so on.

Please note that our classification being mutually inclusive, the words in this category can appear in other categories as well.

**12. Race**: Nepal is a land of racial, ethnic and linguistic diversity. There are more than 125 ethnic groups and languages [18] in the country. *Bahun*, *Chhetri*, *Tamang*, *Magar*, *Newar*, *Gurung*, *Kami*, *Muslim*, *Yadav*, and *Rai* are some of the most populous races. Designating people by their ethnic groups has a long history, starting from religion and followed by rulers, dictators, aristocratic cultures, patricians, and so on. Several Hindu holy scripts categorize people as untouchable [28] (*baishya*, *sudhra*, caste-less etc.) and educated or upper-cast (*Brahmin*, *Chettri* etc.), to avoid eating food touched by lower-class people, and even touching them (e.g. *Damai, Kami le choko khana hudaina*). Traditionally, the upper-cast people are downgraded should they eat the food touched by the lower-cast people or marry them. Due to political awareness and human rights activities, the narratives are changing, and the cast system is slowly fading. But still, racist comments using these terms are highly prevalent in contemporary society. Examples:

- Oi **bhote**, noon bechna chodis? (Hey *bhote*, have you stopped selling salt?)
- **Bahun** tero tuppi ukheldim? (Bahun, can I pluck your tuppi?) *Tuppi – sikha* is a tuft of hair left on the top of shaved head of traditional Hindu; stereotypically associated with Brahmin.
- Yesko buddhi **kami** kaha lagera sojhauna parcha (His head is not right, need to take him to *kami – blacksmith* to straighten it)
- Rajako kaam chhodi **kamiko** dewali (Going to kami's festival over king's assignment) is a traditionally non-tabooed idiom that is considered racist now.

We also found abundant words and phrases that are used to make racist comments. It includes *dhoti, madhise, tapare, pate, chimre, bhote, toppa, pahade, ganaune, kote, sarki, jyami, sinu khane jaat* (carcass eating people), *sikarmi, dakarmi,* and *pode*. The use of these words to a totally unrelated people are offensive outright. Substitution for cast names such as *bahun baje* for *bahun*, *kaji* for *chhetri*, *sahu* for *newar*, *mijar* for *sarki*, and *bishwokarma* for *kami* can serve as euphemisms if they are used properly.

**13. Relation**: Relational offensive terms in Nepali include both human-human relations such as *'maa' chikne* ('mother' fucker) and *randi ko 'chora'* (son of a prostitute) and human-nonhuman relation such as *raandi ko 'baan'* (widow's 'rider'). Other linguistic taboos in this category include *harami* (bastard), *chutiya, rakhel, raandi, batase* (one without a known father) and *rando* which are outright offensive. Contextually offensive terms in this category include *bidhwa, dadha, logne, bau, kuputra, sali,* and *bhena.* Examples:

- **Talai kina chayieo?** *tero baule* banako ho ra? (Why do you need it? did *your father* made it?)
- **Raandi ko chora**, talai pakh na maile janya chu (*Son of widow*, you'll see what I am going to do)

The popular euphemism for a widow is ekal mahila (single women) which also refers an unmarried/divorced woman.

**14. Gender**: Gender-related taboos and offensive terms exist in most of the cultures. *Pothi baseko ramro hudaina, thangne aaimai, gireki aaimai* are historically prevalent sexist phrases in Nepali household. Discrimination based on a person's gender is extremely common in Nepal. Mostly, females and LGBT are found to be affected by this segregation. Several words for making gender bias such as *ghopte, kukhuri, pothi, bakhre, budi kanya, boka, bokie, uttauli, nakharmauli, chothale, maal* (slang referring to a female) are routinely used in offensive remarks. The discriminatory and derogatory use of certain words such as *aaimai, taruni, taruno, bidhuwa* have changed their actual meanings over the course of time. Today, direct use of these words brings discomfort and are mostly avoided or substituted by euphemisms in public use. Additionally, observing thousands of social media comments, we found that a set of words such as *ge, hijada, napubsahk, lesbian* – supposed to refer LGBTQ are misleadingly used to offend others. Examples:

- **Bidhuwa** le *bisait* pari (*Widow* brought a bad luck)
- **Hijada** jastai chhe (She looks like a *transgender*)
- Thangne **aaimai** (Dirty *women*)
- Yo **aaimai** ko jaat lai jati bhane pani lagdaina, jaat janai halchan (A sexist phrase which semantically means "All women are foolish" )

**15. Weapon**: Weapon names are rarely considered offensive when used in general conversation. Our observation shows that they are occasionally used to threaten or attack others. Commonly used terms includes *khukuri* (Nepali knife), *chhura* (small knife), *tarwar* (long knife), *goli* (bullet), *bhala* (spear), and *bikh* (poison). Interestingly, we also observed these words being frequently used on posts related to political disagreements, social criminals, and racist bigotry. Examples:

- **Bhala** le hanna parcha yeslai (Need to strike him with a *spear*)
- Bhukdai gar, chadai **khukuri** khanchas taile (Keep barking, soon you will be struck with a *khukuri – knife*)
- Balatkari lai **jhundyaunu** parcha (We need to *hang* the rapist)

**16. Religion**: There are many religious taboo activities in Nepal such as wearing shoes in temples, touching god's cculpture with the left hand and so on. The religious linguistic taboos in Nepali language are mostly related to death, devil, mystical words (associated with religious texts), and God names. And these taboos are mostly context-dependent e.g. *raakshyas, bhoot, masan, daanab, raawan*. Below are some examples of offensive comments in this category:

- Manav rupi **daanab** hun ini (He is an *evil* in the face of man).
- **Kukarmi daitya**, tero narka ma bas hos (*Immoral evil*, I wish you go to hell)

**17. Geographic Location**: Offending people based on their geographic location or country is a widely discussed topic in Nepal. There are routinely used racial slurs to discriminate people based on their origins. Slurs such as *bihari, indian* and *madhishe* are used for people from Terai (Southern plain) region and the slurs *bhote*[14] and *pahade* are used for people from Pahad (Northern Himalayan and hilly) region. In fact, we encountered hundreds of comments containing these abusive terms in our corpus. Additional, *jhapali, sangjali, jangali, khaire* were also abundant in the corpus. We also identified a few specific locations referring to red light areas such as *ratna park* and *balanju bus park*. Some examples of offensive comments in this category are given below:

- Herdai **ratna-park** ma beluki grahak khojne jasti che ("She looks like the one who seeks for the costumers at *ratna-park* in the evening" in local context, is citing a woman as a whore)
- Thangne **bhoteni** (Dirty *bhoteni*)

**18. Other**: There cannot be a closed set of offensive varieties. Below we list some other interesting varieties that are not previously covered.

**A. Composite Offensive Terms**: Complex offensive terms are formed using the individual offensive terms such as *dhoti ko chhauro* and *muji bhatuwa*. More common case is to use a taboo term with a contextually offensive term such as *randi ko baan, sala kukur, gobar gwatch* and *tapare bahun*. Although these terms mostly collocate as bigrams, they might appear at different locations in a sentence. In the sentence "bahun jati sabai *tapare* nai bhaye", the composite term *tapare bahun* is formed using *bahun* and *tapare* appearing at different locations.

In general, the degree of offensiveness increases with the increase in offensive terms count. For instance, the expression *muji bhatuwa* is more offensive than the individual offensive terms *muji* and *bhatuwa*.

**B. Imported Terms**: There is a large list of imported offensive terms from other local and foreign languages such as Hindi (*dekha jayega, murga, ulluka patta, behen chod, bhosadi*), Arabic (*aaulad, saitan*) and English (*over, virgin, blue, bra, saiko – psycho, fuck*).

**C. Code Switching**: Code switching is a process of forming a composite term by mixing more than one language. Examples of

---
[14]Bhote, also known as Bhotiya are group of people related with Tibetan ethnically or linguistically

offensive terms due to code switching include *over nabol* (English + Nepali), *randiko aaulad* (Nepali + Arabic), and *ta murga* (Nepali + Hindi)

**D. Exclamation**: Also known as *Bismayadibodhak* in Nepali, exclamations in communication pertain to strong emotions, especially anger, surprise, dissatisfaction, or agony. Words such as *thukka, thuiya, chyaa, murdabad, jindabaad, thu-thu, bajo paros, mari ja* etc. are a few examples of exclamatory terms in Nepali that can appear in offensive comments.

## 5 EUPHEMISMS IN NEPALI

Sometimes the use of taboo terms is unavoidable, e.g., patients describing doctors about their genital-related issues, talking with kids about bodily excretions, mentioning disabled persons in a conversation, and so on. Euphemisms are the ways people use to lessen or completely remove the unpleasantness of taboo expressions. They are often the replacement words or phrases that substitute the offensive terms and expressions to neutralize or reduce the nuisance [10]. For example, in English the phrase *make love* is preferred over *fuck* in a conversation with mixed participants [15, 17]. Similarly, instead of using *underdeveloped nations* for Third World countries, the United Nations calls them *less-developed countries* [10]. In this section, we discuss common types of euphemisms used in Nepali. Many of these euphemisms are applicable to the offense varieties we listed in Section 4.

(1) *Synonym*: Synonyms of a term can help to eliminate the offensiveness of a taboo term. The non-offensive terms *rajkhani, pisab* and *disha* are the synonyms for taboo terms *geda* (testis), *muut* and *gu* respectively.

(2) *Word Replacement*: The changes in sociopolitical structure and human right movements have helped to coin and encouraged to use non-offensive terms instead of tabooed terms: *Ykal mahila* for *bidhuwa*, *Jestha Nagarik* for *buda-budi*, *Tesro Lingi* for *Hijada*, *Sadak balbalika* for *Khate* and *Yaunkarmi* for *Besya* etc.

(3) *Circumlocution*: Circumlocution is an evasive way of expression using multiple words rather than being explicit with fewer and apter words. It is commonly practiced to avoid the use of taboo terms. One popular way of circumlocution is to use the definitions of offensive terms. Examples include *Pisap ferne anga / Pisap aaune thaau* instead of *laado or turi* for man and *puti* for women.

(4) *Word Translation*: Sometimes translation to another language can enfeeble the offensiveness of a taboo term. In Nepali, the translation is typically done in English or Sanskrit for euphemism. It is easier to use *mens or period* instead of *nachhune* menstruation. Similarly, *stool* is commonly preferred over *gu* in hospitals. Occasionally, the translated forms are shorter than the offensive terms and thus are preferred. One such example is *nipple* instead of *dudh ko tuppo*.

(5) *Child Language*: People in any society have different terms that replace the corresponding tabooed terms when talking with a child. Here are few examples in Nepali: *aaka for gu, susu* for *mut*, *tun-turi* or simply *turi* for *lado* (male genital), *hawa* for *paad* etc.

(6) *Suffix*: Addressing a person with his first name is very common in western societies. In Nepali culture, which has an honorific system, addressing people having more honor (e.g. parents) with their names is considered offensive. Use of person names, however, is common among the peers or to the persons with lower honors, e.g., children. There are some suffixes (e.g. *jee* and *jyu*) that can be attached to person names to increase the politeness. For example, using Ramjee for Ram and Sitajee for Sita in a conversation can provide more honor than using the direct names. Adding *jee* after a person's name can also mean the speaker has a good relationship with the person and therefore is common across friends. Also, *jee* can be used after the surnames (e.g. Dahaljee) or nicknames (e.g. Prachanda jee).

Use of *jyu* provides respects to superiors, e.g., ministers, presidents and managers. Unlike the suffix *jee*, this suffix is mostly used after the complete name of a person e.g. Madhav Nepal *jyu*. It is also commonly used in formal writing, official letters, legalese and public speech.

(7) *Family Alias*: Using alias for names inside a family is a common practice in Nepali culture. Instead of using husband's name directly (which is considered abnormal), a wife uses aliases such as "X ko Baba" (X is offsprings' name), baba, and daddy. Similarly, aliases for wife include sani, kanchhi, and "X ki ama". However, parents can directly use the first names of their children. Also, it is common to use rank alias for children e.g. jetho/jethi (the first male/female child), mailo/maili(the second male/female child), kancho/kanchhi (the last male/female child), etc.

(8) *Relation Name*: This is probably the most widely used approach in Nepali culture to avoid using person name in a conversation. Relations to superiors in the family tree *(hajurba/hajurama, buwa/ama)*, elder siblings *(daju/didi)*, younger siblings *(bhai/bahini)*, rank of siblings by order *(jetha/jethi, maila/maili, saila/saili, kaila/kaili, antare/antari, thaila/thaili, jantare/jantarni, kanchha/kanchhi)*, peers *(sathi, mitra)*, inferiors *(chhora/chhori, bhanja/bhanjee, keta/keti)* and other alias *(sane/sani, maya, mayalu, maiya, mitjyu, mitini)* etc.

(9) *Title*: People of high political rank or that are of a prominent post of an organization are addressed using their titles or titles followed by their names. Examples include *Pradhan Mantri* (Prime Minister), *Upapradhan mantri* (Vice-president), *Rastrapati* (President of a country), *Sabhamukh* (House Speaker), *Adyakchya* (President), *Swastha Mantrin* (Health Minister), *Purba Mantri* (Ex-minister), etc. Adding "Sir" after the name is also a popular practice e.g. Hari sir, Madhav sir, etc.

(10) *Metaphor*: Metaphoric expressions are also used as euphemisms. If someone is very talkative (negative connotation), he is called *Narad*, a Vedic sage (in Hindu religion) who is a traveling musician, storyteller, news carries, and enlightening wisdom. Other examples include: *dhunga jasto mutu* (heart is like a stone) for *nisthuri* (cruel), *Gobar Ganesh* for *pata murkha* (stupid).

(11) *Idiom*: Idioms are very common in both written and spoken languages. Some examples in this category include: *sangai*

*uth bas garnu* (up down together) for *sex*, *para sarnu* or *nachune hunu* for *menstruation*, *swarga ma baas hunu* (move to heaven) for *death*, and *chura phutnu* (to get bangle broken) for *bidhuwa hunu* (to be a widow).

(12) *Proverb*: Proverbs can be used as euphemisms. One such example is *bhukne kukurle tokdaina* (barking dog seldom bites) which can contextually mean *bataure* (braggart or blowhard).

Interestingly, euphemisms more than needed can reverse the purpose of euphemism [10]. The expression *"Yaauna karmi ko pisab pherne anga"* has a negative meaning although euphemistic terms *Yaauna karmi* and *pisab pherne anga* are used in the expression.

## 6 CONCLUSION AND FUTURE WORK

Social media are excellent platforms for sharing information. They are also misused to spread hate speech through offensive posts and comments. With enormous amount of content generated every day, automated methods are required to prevent them from spreading. Because offensiveness is a socio-cultural and situational phenomenon, language agnostic automated methods do not work for hate speech detection. Language specific study is necessary to understand what can be considered offensive. To this end, we presented our corpus-based study of offensive language in Nepali. We primarily focused on linguistic taboos and euphemisms. We believe this is the first study of its kind.

We collected and studied over 7000 social media posts containing high degree of linguistic taboos and offensive terms. Furthermore, we carefully extracted and characterized over 1000 tabooed and contextually offensive terms into 18 different varieties. The annotated data set is made publicly available at Github[15]. Besides the universal taboo topics like Sex, Body Part, and Excretion, we identified many varieties that are unique to Nepali language and culture e.g. Relation and Plant Names. Since no related prior work or resource exists for Nepali, the study and the data set lay a foundation for future research on this important topic. For example, in hate speech detection task, a baseline system can be built from the aforementioned data set that can flag content (e.g. a Tweet or Reddit post) as offensive if the terms in the data set appear in the content. It can further help to do a fine grained classification of linguistic offenses.

We also studied euphemisms – ways for people to lessen or completely remove the offensiveness. We discussed 12 common types of euphemisms in Nepali. While euphemisms such as Circumlocution and Child Language are common across the cultures in the world, many other discovered euphemisms are specific and unique to Nepali culture.

Offensive language detection in Nepali is the immediate next topic we are considering to fight against racism, discrimination and cyberbullying. Other future work includes constructing an ontology of offensive categories. For example, many of the identified varieties such as Disability, Race, Religion, Gender, and Location can be put together under a higher-order *Discriminatory* category. Likewise, Plant and Animal can be grouped, and so on. The study of behavioral taboos is another future topic in this direction.

---

[15] https://github.com/nowalab/offensive-nepali

## ACKNOWLEDGMENTS

We are grateful to Professor Dr. Kumar Prasad Koirala and Mr. Daniel Whyatt for their valuable inputs and inspiration. We would also like to acknowledge Nepali Shabdakosh (http://nepalishabdakosh.com) for providing synonyms of the offensive terms.